# Modelling resource allocation in uncertain system environment through deep reinforcement learning

**Neel Gandhi**
Student, School of Technology
Pandit Deendayal Energy University
Gandhinagar, Gujarat
`neel.gict18@sot.pdpu.ac.in`

**Shakti Mishra**
Associate Professor, School of Technology
Pandit Deendayal Energy University
Gandhinagar, Gujarat
`shakti.mishra@sot.pdpu.ac.in`

## ABSTRACT

Reinforcement Learning has applications in field of mechatronics, robotics, and other resource-constrained control system. Problem of resource allocation is primarily solved using traditional predefined techniques and modern deep learning methods. The drawback of predefined and most deep learning methods for resource allocation is failing to meet the requirements in cases of uncertain system environment. We can approach problem of resource allocation in uncertain system environment alongside following certain criteria using deep reinforcement learning. Also, reinforcement learning has ability for adapting to new uncertain environment for prolonged period of time. The paper provides a detailed comparative analysis on various deep reinforcement learning methods by applying different components to modify architecture of reinforcement learning with use of noisy layers, prioritized replay, bagging, duelling networks, and other related combination to obtain improvement in terms of performance and reduction of computational cost. The paper identifies problem of resource allocation in uncertain environment could be effectively solved using Noisy Bagging duelling double deep Q network achieving efficiency of 97.7% by maximizing reward with significant exploration in given simulated environment for resource allocation.

*Keywords* Reinforcement Learning · Resource Allocation · Noisy Bagging D3 Q Network · Control Robotic systems

## 1 Introduction

Deep reinforcement learning [1] has intensively been used to solve complex problems that are rather difficult to be solved using conventional techniques. Reinforcement learning aims at achieving long term reward by the mechanism of agent interacting with the environment and receiving an appropriate reward through a mechanism of state transition and evaluative feedback. Deep reinforcement learning, in recent years, has found applications in the field of robotics [2], system control [3] and other resource allocation problems [4–6]. Deep reinforcement learning is generally used in cases of exploring an unknown environment where an agent learns by interacting with the environment and develops a suitable policy aiming at maximizing long-term reward. Reinforcement learning maintains balance between exploration and exploitation to obtain optimal result [7]. Also, new horizons of decision making have been created using the power of reinforcement learning that were earlier not possible using supervised machine learning methods. Reinforcement learning is used in resource Management that was found helpful in cases of job scheduling in computer clusters [8], relay selection in internet telephonic [9], traffic congestion control [10], adaptation of bit rate in video streaming [11] and other variants of reinforcement learning methods have been adopted for resource allocation in field of games [5] and business process management [6]. This paper uses different variants of deep reinforcement learning algorithms majorly based on Q learning approaches for the purpose of resource allocation. The paper suggests tools and techniques for solving the problem of resource allocation using reinforcement learning. The paper contributes towards development of reinforcement learning models for resource allocation in control systems and robotics. The paper is divided into various section as follows:-

Section 1 includes application of Deep Reinforcement Learning in resource allocation problems and their respective approaches in field of robotics, control system, mechatronics, and other related fields followed by Problem statement formulation along with description of the proposed Noisy Bagging D3 (duelling double deep )Q learning method is provided along with an algorithmic view in Section 2.Further, results are obtained from various RL algorithms and comparison of RL algorithm is done against our Noisy Bagging D3 (duelling double deep )Q learning with respect to exploration versus reward criteria in Section 3. Also, resource utilization graphs are provided in order to understand efficiency of the system. Section 4 includes concluding remarks for our proposed Noisy Bagging D3 Q learning method in resource allocation problem in an uncertain environment along with future perspective are provided.

## 2 Applications of Reinforcement learning model in resource allocation

Reinforcement learning has found applications in Static and dynamic type of resource allocation from past couple of year [4, 12].Researchers have used open loop and closed loop systems and check their performance using model based policies and result show significant performance in terms of reinforcement learning algorithm. Vengerov [13] proposed reinforcement learning along with fuzzy rule based approach for the purpose of dynamic resource allocation.Resource allocation has found applications in robotics [14].Reinforcement learning could be applied in the form of of Mobile Edge computing [15–17] were major research is done by various scientist for resources allocation at network edge.Also,research has been done in field of cloud and power management in robotics were agent has to identify a control policy for an interaction with partially observable system at the same time providing management capabilities in Complex environment [14, 18].Many Framework have also been developed for cloud Resource Management,some of which include deep reinforcement learning algorithm in their internal configuration. [19] Resource Management [12] and business process management have also utilise deep reinforcement learning approaches.They have been utilised for control of vehicles through process of control resource with aid of internet of things.Reinforcement learning has been used for applications of UAV and vehicle Resource Management along with the integration of internet of things in various applications [14, 20].Also,grid computing [21],task scheduling and other allocation task are being handled by dynamic resource allocation algorithms derived from reinforcement learning.QCF algorithm was developed by Lian et al [22] that integrated Q learning and chain feedback was used for resource allocation having a significant high convergence.In this paper,we have proposed algorithm of noisy Bagging D3 learning to suitable for given allocation problem after experimenting and analysing various Deep Q learning approaches.

## 3 Proposed Methodology

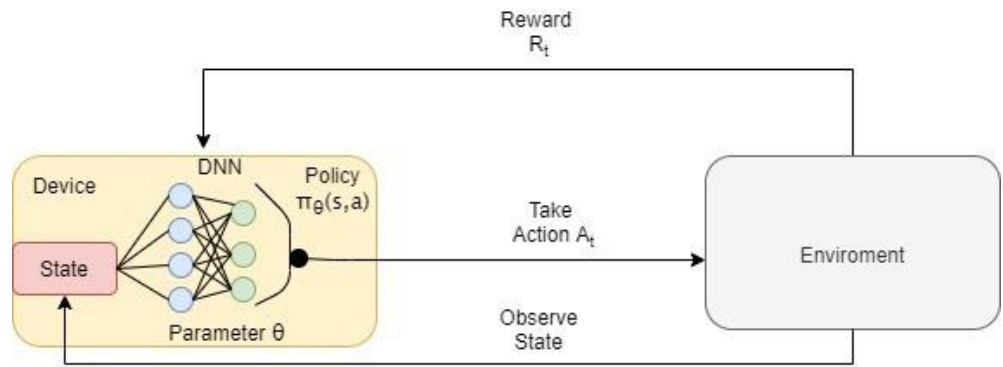

Figure 1: *Learning Mechanism in Reinforcement Learning*

Reinforcement learning applied to resource allocation problems where the agent reacts with the environment and learns by making and rectifying errors [23]. For example, in the case of resource allocation, if a particular item is allocated to a particular resource that is incompatible, then it will immediately ask for a change request. In this way, a reinforcement learning agent reacts with the environment, gets a particular reward, and stores observation for future time steps to achieve maximum cumulative long-term reward. Reinforcement learning is helpful in solving our complicated and complex problem of resource allocation that is changing with passage of time and it proves to be superior in an uncertain environment rather than the unsupervised and supervised learning approaches. The proposed model in the long-run performs similar to human-level intelligence. Also, errors are corrected by the model with the help of reinforcement learning algorithms. Reinforcement learning algorithms prove to be effective for the purpose of robotic applications such as learning how to walk, locate a particular object in an unknown environment or resource allocation. The advantage of using reinforcement learning for resource allocation problems is the property of adaptability to any type of setting as it is bound to learn from its experience even in absence of training data set. The proposed noisy bagging D3 reinforcement learning methods prove to be superior compared to any other algorithms. Noisy layer helps us to solve the exploration and exploitation dilemma indeed to maximize the long-term reward. The mechanism of functioning of reinforcement learning algorithm is depicted by fig1.

### 3.1 Problem Formulation

A system has a well defined number of resources and variable number of items varying in accordance to time. The goal of reinforcement learning agents is assignment of resources to items in optimal manner following policy selecting actions with greatest Q value. Assumption is made that each item would require one resource at a time and would be able to hold that for at least a few periods of time, this type of setting is essential in case of robotic systems where a particular operation has to be performed using a specific resource. Task of reinforcement learning algorithm is to assign resources to items in optimal manner to minimize changes required frequently alongside avoid cases of unused resources and non-performing items in the system environment. An imaginary instance of a simulation environment has been depicted in fig.2 to get an illusion about the actual problem statement. Various reinforcement learning algorithms were applied to given problems in order to maximize the long-term reward and try to minimize the loss in case of improper resource allocation. Simulation environment can be depicted in fig. 2

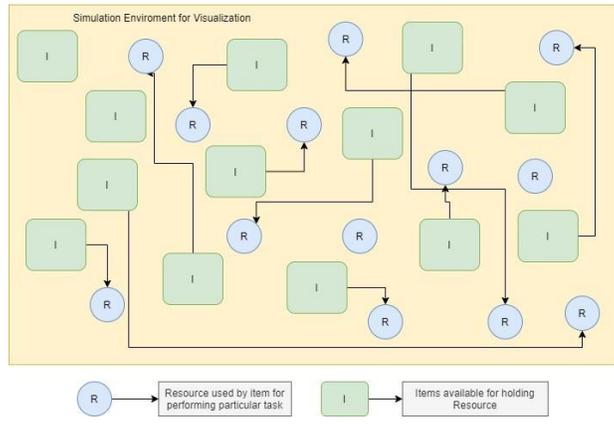

Figure 2: *Simulation Environment*

#### 3.1.1 State space, Action space and Rewards

Here, detailed description regarding state space, action space and reward function of our problem formulated is described as follows:

**State space-** It helps us to understand current allocation of resources and resources available for being allocated to different items. In our environment, we have taken a fixed state representation in order to input into our neural network. Suppose, there are M resources each being allocated to a single item according to requirement. The task of reinforcement learning algorithms would be to change the state of the system by allocating resources. Items have the advantage of being constrained from maximum space due to a fixed set of resources and items make the learning process more efficient. Also, we have allocated one resource per item and allocated some delay before allocating it to another resource in case of a change request. The simulation also provided many items at a certain time stamp in order to inspect the performance of the algorithm.

**Action space-** Considering M resources for given number of items, for task of resource allocation, Several ways are possible for task of resource allocation generating large action space of $2^M$, which would make the performance of reinforcement learning more challenging and time consuming. So, agents are allowed to take more than one action at a time and can allocate resources to items. Also, in case of a change request by item, it requires some delay before allocation of another resource. Agent observes the state of the system and takes appropriate actions in order to reduce the number of re-allocation of resources and improve efficiency of the system trying to reduce the number of changes required at the same time being efficient in case of utilization of resources.

**Rewards-** Using reinforcement learning algorithms, agents try to maximize cumulative long-term reward. In the given problem statement, the maximum reward that could be achieved is zero as it is a continuous process without definite terminal state due to uncertain environment simulation settings. Reinforcement learning algorithms try to minimize loss as long as possible to maximize reward in our resource allocation problem. The equation for reward could be illustrated as

$$reward = -abs(unutilized\ resource - target\ unutilized\ resource)$$

### 3.2 Proposed Reinforcement Learning Model

Deep Q networks aim to increase the future rewards with the help of selecting the greatest Q among available sets of actions using a particular policy for the state transition. Q learning follows a Bellman equation.

$$Q(s, a) = r + \gamma \cdot \max Q(s', a') \tag{1}$$

where Q(s,a) is a resultant of reward received in addition to discounted maximum Q value of the next action and gamma() represents discount rate. Double deep Q networks [24] is used for the purpose of improving network architecture where current state and target Q are compared with input features using policy net. In addition to policy net,target net is introduced in order to reduce the number of operations due to prediction of Q values. Policy network replicated itself across a network every thousand steps for further improvement in performance. For improvement of policy net, Duelling Deep Q networks [25] were introduced where policy net was split into two components resulting into one component being the state value of the model and other being an advantage in difference of Q between different actions taken by learning model for resource allocation.Duelling Double Deep Q networks were used for achieving optimal policy net along with advantage of target network. Aggregation was used for producing actions for resource allocation. In order to deal with exploration-exploitation dilemma and find a better alternative to traditional epsilon greedy exploration to improve performance of the model in an uncertain environment, noisy layers [26] were introduced where factorized Gaussian Noise was preferred over Independent Gaussian Noise. Factorized Gaussian Noise comprised of two vectors, first vector for length of input sequence and second vector for length of output sequence, further specify mathematical function is used for calculating matrix multiplication and predict outcome.It results in formation of randomized matrix that can be then added to our model in case of uncertain environment. Noisy layers were found appropriate for our model to get rid of Epsilon greedy exploration and to effectively deal with the exploitation exploration dilemma in an uncertain environment. Prioritized replay [27] didn't prove to be effective in an uncertain environment due to greater attention towards wrong predictions to minimize the loss but it didn't prove to be effective due to uncertainty in the environment and other related factors. Lastly, bootstrap aggregation or bagging [28] was used to train our model where different samples trained on the same memory having difference in starting weights as well as trained on different bootstrapped networks from that memory. Actions were taken in accordance with random choice or most commonly recommended action by aggregator. Bagging was helpful in exploration due to suggestions of different suggested actions from different networks.

The combination of above methodologies were integrated and applied to our proposed noisy bootstrap aggregation duelling double deep Q network in uncertain environment were actions are taken in accordance to particular policy trying to maximize reward in our model based environment with use of policy based methods.The model tried to maximize the cumulative long term reward in case of of resource allocation.The steps followed by the model included:-

1) Simulation environment was created,memory was set up,selection of policy net,target net and setting up noisy layer is done in the initial step of our resource allocation simulation

2) Bootstrap samples are derived from memory and trained to recommend most appropriate action.

3) Inside different nets training was done to suggest next action in order to maximize reward report.Aggregator is used for selection of best possible action.Further,storing of the state,next state,reward and terminal to the memory operation is performed after choosing best recommended action from among different samples

4) Lastly,evaluation was done for performance of proposed Noisy bagged D3 reinforcement learning algorithm

**Noisy bootstrapped aggregation Bagging D3 Q learning** approach proves to be superior compared to other variants of Q learning models for resource allocation problems. The noisy layers are effective in exploration of uncertain environments and act as an effective alternative to epsilon-greedy exploration. Duelling nature helps in getting the state value as well as difference in Q value between different actions, which is helpful in selection of appropriate action with greatest Q value. Bootstrap Aggregation(Bagging) helps in measuring the uncertainty of action which is effective in selecting the best possible actions from different networks based on criteria of random selection or average selections or most recommended action. Bagging was also found to be helpful in aiding stages of exploration where networks provide different suggested actions. Hence, Noisy bootstrapped aggregation ('Bagging') D3 Q learning is effective in terms of reward as well as efficiency for resource allocation problem in uncertain environment.

Algorithm implementation of Reinforcement Learning for dealing with problem of resource allocation is depicted by underlying algorithm view

---

**Algorithm 1** An overview of steps involved in resource allocation using RL algorithm

Create simulation environment
Create memory
Create policy net
Create target net

**while** *Training episodes not complete* **do**
    Reset simulation
    **while** *not in terminal state* **do**
        Recommend action from policy net
        Pass action to simulation
        Increase time-step in simulation
        Receive (next state, reward, terminal,info) from simulation
        Store (state, next state, reward, terminal)to memory
        Update policy net
    **end**
    Update target net
**end**
Evaluation for performance of policy net

---

## 4 Results

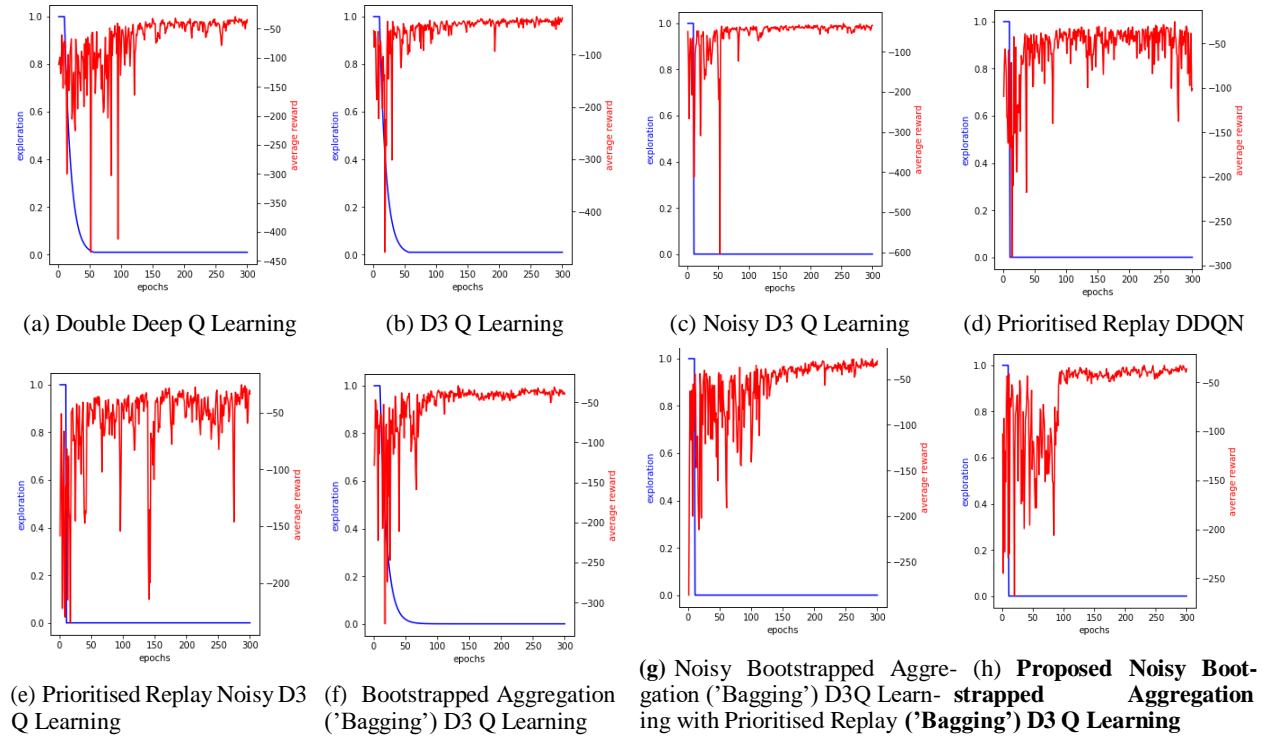

Figure 3: Exploration vs Reward

(a) Double Deep Q Learning
(b) D3 Q Learning
(c) Noisy D3 Q Learning
(d) Prioritised Replay DDQN
(e) Prioritised Replay Noisy D3 Q Learning
(f) Bootstrapped Aggregation ('Bagging') D3 Q Learning
(g) Noisy Bootstrapped Aggregation ('Bagging') D3Q Learning with Prioritised Replay
**(h) Proposed Noisy Bootstrapped Aggregation ('Bagging') D3 Q Learning**

The result obtained after simulation choosing various types of reinforcement learning algorithms right from double Q network to noisy bootstrap aggregation duelling double Q network with prioritized replay showed that noisy bootstrap aggregation D3 Q learning approach was supposed to be the best algorithm in terms of efficiency for resource allocation. Hence, effective for operation of resource allocation in robotic control systems. Few algorithms having prioritized replay suffered from under capacity for prolonged amount of time with significantly less accuracy due to reason uncertain/fuzzy environment of simulation. Also, noisy layers proved to effective in improving accuracy of the system. Duelling nature, in some cases, would increase or decrease efficiency of system in the environment due to splitting of policy net. Lastly, we have identified that noisy bagging D3 Q network approach was found to be superior in terms of exploration versus reward illustrated by fig 3 and also utilization of resources illustrated by fig 4. Noisy bagging D3 Q network shows that it is the best suitable algorithm for resource allocation. Comparative analysis of various algorithms on basis of reward and resource utilization by items or agent present in the system along with their respective calculation for analyzing efficiency of different reinforcement learning algorithm is illustrated by table given below

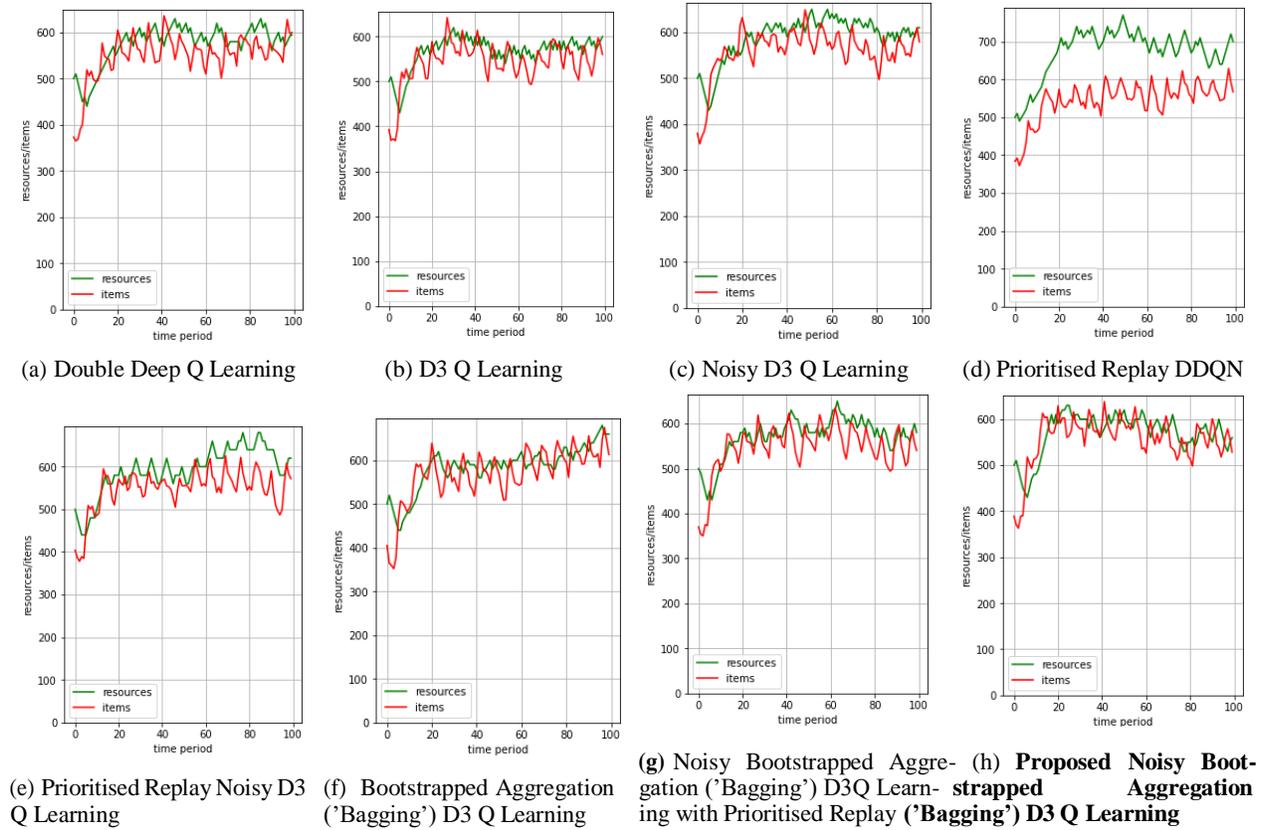

(a) Double Deep Q Learning  
(b) D3 Q Learning  
(c) Noisy D3 Q Learning  
(d) Prioritised Replay DDQN  
(e) Prioritised Replay Noisy D3 Q Learning  
(f) Bootstrapped Aggregation ('Bagging') D3 Q Learning  
(g) Noisy Bootstrapped Aggregation ('Bagging') D3Q Learning with Prioritised Replay  
(h) **Proposed Noisy Bootstrapped Aggregation ('Bagging') D3 Q Learning**

Figure 4: *Resource utilization graphs*

| ① | Double Deep Q Learning |
|---|---|
| ② | D3 Q Learning |
| ③ | Noisy D3 Q Learning |
| ④ | Prioritised Replay DDQN |
| ⑤ | Prioritised Replay Noisy D3 Q Learning |
| ⑥ | Bootstrapped Aggregation ('Bagging') D3 Q Learning |
| ⑦ | Noisy Bootstrapped Aggregation ('Bagging') |
| ⑧ | **Proposed Noisy Bootstrapped Aggregation ('Bagging') D3 Q Learning** |

|  | ① | ② | ③ | ④ | ⑤ | ⑥ | ⑦ | ⑧ |
|---|---|---|---|---|---|---|---|---|
| time period under capacity | 28 | 22 | 23 | 0 | 13 | 38 | 27 | **38** |
| time period over capacity | 72 | 77 | 77 | 100 | 86 | 62 | 73 | **62** |
| average items performing | 550 | 543 | 560 | 545 | 548 | 570 | 546 | **557** |
| average resources utilized | 575 | 567 | 589 | 673 | 591 | 584 | 571 | **570** |
| % efficiency | 95.6 | 95.8 | 95 | 81 | 92.7 | 97.5 | 95.7 | **97.7** |

## 5 Conclusion

Reinforcement learning models have been used for resource allocation in robotics system to take control decisions with significant performance at runtime. We propose a reinforcement learning method of noisy bagging D3 Q network for modelling resource allocation in a way to maximize long-term reward, balance exploration-exploitation dilemma and performing operations with significant efficiency in the long run. The proposed method could be applied to other resource allocation problems like task scheduling, job scheduling and similar complicated scheduling tasks in the field of robotics, mechatronics and other control systems. The proposed model proved to be feasible and efficient at the same time being effective in taking control decisions for resource allocation in an uncertain environment. In order to obtain maximum performance and solve a complicated problem in control systems. We have also compared our proposed method with different variants of deep reinforcement learning models across various evaluation parameters. Reinforcement learning methods applied for resource allocation tasks still face few major challenges like explainability, scalability, complexity and flexibility. For future works, we can develop a generic framework for different kinds of allocation problems in resource constrained environments. The paper contributes towards the field of artificial intelligence by virtue of introduction of automation in dynamic resource allocation for robotic control systems.

## References

[1] Maxim Lapan. *Deep Reinforcement Learning Hands-On: Apply modern RL methods, with deep Q-networks, value iteration, policy gradients, TRPO, AlphaGo Zero and more*. Packt Publishing Ltd, 2018.

[2] Jens Kober, J Andrew Bagnell, and Jan Peters. Reinforcement learning in robotics: A survey. *The International Journal of Robotics Research*, 32(11):1238–1274, 2013.

[3] Tianshu Chu, Sandeep Chinchali, and Sachin Katti. Multi-agent reinforcement learning for networked system control. *arXiv preprint arXiv:2004.01339*, 2020.

[4] Gerald Tesauro, Nicholas K Jong, Rajarshi Das, and Mohamed N Bennani. A hybrid reinforcement learning approach to autonomic resource allocation. In *2006 IEEE International Conference on Autonomic Computing*, pages 65–73. IEEE, 2006.

[5] Si-Ping Zhang, Jia-Qi Dong, Li Liu, Zi-Gang Huang, Liang Huang, and Ying-Cheng Lai. Reinforcement learning meets minority game: Toward optimal resource allocation. *Physical Review E*, 99(3):32302, 2019.

[6] Zhengxing Huang, Wil M P van der Aalst, Xudong Lu, and Huilong Duan. Reinforcement learning based resource allocation in business process management. *Data Knowledge Engineering*, 70(1):127–145, 2011.

[7] Ithan Moreira, Javier Rivas, Francisco Cruz, Richard Dazeley, Angel Ayala, and Bruno Fernandes. Deep Reinforcement Learning with Interactive Feedback in a Human–Robot Environment. *Applied Sciences*, 10(16):5574, 2020.

[8] Robert Grandl, Ganesh Ananthanarayanan, Srikanth Kandula, Sriram Rao, and Aditya Akella. Multi-resource packing for cluster schedulers. *ACM SIGCOMM Computer Communication Review*, 44(4):455–466, 2014.

[9] Junchen Jiang, Rajdeep Das, Ganesh Ananthanarayanan, Philip A Chou, Venkata Padmanabhan, Vyas Sekar, Esbjorn Dominique, Marcin Goliszewski, Dalibor Kukoleca, and Renat Vafin. Via: Improving internet telephony call quality using predictive relay selection. In *Proceedings of the 2016 ACM SIGCOMM Conference*, pages 286–299, 2016.

[10] Keith Winstein and Hari Balakrishnan. Tcp ex machina: Computer-generated congestion control. *ACM SIGCOMM Computer Communication Review*, 43(4):123–134, 2013.

[11] Yi Sun, Xiaoqi Yin, Junchen Jiang, Vyas Sekar, Fuyuan Lin, Nanshu Wang, Tao Liu, and Bruno Sinopoli. CS2P: Improving video bitrate selection and adaptation with data-driven throughput prediction. In *Proceedings of the 2016 ACM SIGCOMM Conference*, pages 272–285, 2016.

[12] Hongzi Mao, Mohammad Alizadeh, Ishai Menache, and Srikanth Kandula. Resource management with deep reinforcement learning. In *Proceedings of the 15th ACM workshop on hot topics in networks*, pages 50–56, 2016.

[13] David Vengerov. A reinforcement learning approach to dynamic resource allocation. *Engineering Applications of Artificial Intelligence*, 20(3):383–390, 2007.

[14] Yi Liu, Huimin Yu, Shengli Xie, and Yan Zhang. Deep reinforcement learning for offloading and resource allocation in vehicle edge computing and networks. *IEEE Transactions on Vehicular Technology*, 68(11):11158–11168, 2019.

[15] Ji Li, Hui Gao, Tiejun Lv, and Yueming Lu. Deep reinforcement learning based computation offloading and resource allocation for MEC. In *2018 IEEE Wireless Communications and Networking Conference (WCNC)*, pages 1–6. IEEE, 2018.

[16] Zehui Xiong, Yang Zhang, Dusit Niyato, Ruilong Deng, Ping Wang, and Li-Chun Wang. Deep reinforcement learning for mobile 5G and beyond: Fundamentals, applications, and challenges. *IEEE Vehicular Technology Magazine*, 14(2):44–52, 2019.

[17] Nguyen Cong Luong, Dinh Thai Hoang, Shimin Gong, Dusit Niyato, Ping Wang, Ying-Chang Liang, and Dong In Kim. Applications of deep reinforcement learning in communications and networking: A survey. *IEEE Communications Surveys Tutorials*, 21(4):3133–3174, 2019.

[18] Yue Jin, Makram Bouzid, Dimitre Kostadinov, and Armen Aghasaryan. Model-free resource management of cloud-based applications using reinforcement learning. In *2018 21st Conference on Innovation in Clouds, Internet and Networks and Workshops (ICIN)*, pages 1–6. IEEE, 2018.

[19] Shuiguang Deng, Zhengzhe Xiang, Peng Zhao, Javid Taheri, Honghao Gao, Jianwei Yin, and Albert Y Zomaya. Dynamical resource allocation in edge for trustable Internet-of-Things systems: A reinforcement learning method. *IEEE Transactions on Industrial Informatics*, 16(9):6103–6113, 2020.

[20] Jingzhi Hu, Hongliang Zhang, Lingyang Song, Zhu Han, and H Vincent Poor. Reinforcement learning for a cellular internet of UAVs: Protocol design, trajectory control, and resource management. *IEEE Wireless Communications*, 27(1):116–123, 2020.

[21] Jun Wu, Xin Xu, Pengcheng Zhang, and Chunming Liu. A novel multi-agent reinforcement learning approach for job scheduling in grid computing. *Future Generation Computer Systems*, 27(5):430–439, 2011.

[22] Chuanqiang LIAN, Xin XU, Jun WU, and Zhaobin LI. Q-CF multi-Agent reinforcement learning for resource allocation problems. *CAAI Transactions on Intelligent Systems*, 6(2):95–100, 2011.

[23] Todd Hester, Matej Vecerik, Olivier Pietquin, Marc Lanctot, Tom Schaul, Bilal Piot, Andrew Sendonaris, Gabriel Dulac-Arnold, Ian Osband, and John Agapiou. Learning from demonstrations for real world reinforcement learning. 2017.

[24] Hado Van Hasselt, Arthur Guez, and David Silver. Deep reinforcement learning with double q-learning. In *Proceedings of the AAAI Conference on Artificial Intelligence*, volume 30, 2016.

[25] Ziyu Wang, Tom Schaul, Matteo Hessel, Hado Hasselt, Marc Lanctot, and Nando Freitas. Dueling network architectures for deep reinforcement learning. In *International conference on machine learning*, pages 1995–2003. PMLR, 2016.

[26] Meire Fortunato, Mohammad Gheshlaghi Azar, Bilal Piot, Jacob Menick, Ian Osband, Alex Graves, Vlad Mnih, Remi Munos, Demis Hassabis, and Olivier Pietquin. Noisy networks for exploration. *arXiv preprint arXiv:1706.10295*, 2017.

[27] Tom Schaul, John Quan, Ioannis Antonoglou, and David Silver. Prioritized experience replay. *arXiv preprint arXiv:1511.05952*, 2015.

[28] Ian Osband, Charles Blundell, Alexander Pritzel, and Benjamin Van Roy. Deep exploration via bootstrapped DQN. *arXiv preprint arXiv:1602.04621*, 2016.